# Do Large Language Models Need Intent? Revisiting Response Generation Strategies for Service Assistant

Inbal Bolshinsky
School of Computer Science,
Faculty of Sciences
Holon Institute of
Technology

Shani Kupiec
School of Computer Science,
Faculty of Sciences
Holon Institute of
Technology

Almog Sasson
School of Computer Science,
Faculty of Sciences
Holon Institute of
Technology

Yehudit Aperstein
Intelligent Systems,
Afeka Academic College of
Engineering
Tel Aviv Israel

Alexander Apartsin
School of Computer Science,
Faculty of Sciences
Holon Institute of
Technology

*Abstract*—In the era of conversational AI, generating accurate and contextually appropriate service responses remains a critical challenge. A central question remains: Is explicit intent recognition a prerequisite for generating high-quality service responses, or can models bypass this step and produce effective replies directly? This paper conducts a rigorous comparative study to address this fundamental design dilemma. Leveraging two publicly available service interaction datasets, we benchmark several state-of-the-art language models, including a fine-tuned T5 variant, across both paradigms: Intent-First Response Generation and Direct Response Generation. Evaluation metrics encompass both linguistic quality and task success rates, revealing surprising insights into the necessity or redundancy of explicit intent modelling. Our findings challenge conventional assumptions in conversational AI pipelines, offering actionable guidelines for designing more efficient and effective response generation systems.

## I. INTRODUCTION

The rapid advancement of large language models (LLMs) has revolutionized the development of conversational agents and automated service systems. These systems are increasingly capable of handling complex user interactions, ranging from simple information retrieval to sophisticated transactional dialogues. At the heart of these systems lies a critical design decision: should the agent explicitly identify the user's intent before generating a response, or can it directly produce an appropriate response without this intermediate step? Despite the practical implications of this question, a lack of systematic investigation remains into how this choice affects response quality, accuracy, and system efficiency.

Traditionally, dialog systems have relied on a pipeline architecture where intent classification precedes response generation. This sequential approach aligns with human reasoning processes, allowing for modularity and interpretability. By first determining the user's intent, such as a request for account balance information or a complaint about service quality, the system can more accurately retrieve or construct a relevant response. However, this design introduces additional computational overhead and depends on the accuracy of the intent classification component. Errors at this stage can cascade, leading to inappropriate or irrelevant responses.

With the emergence of powerful generative models such as GPT-family models and encoder-decoder architectures like T5, a new paradigm has emerged: Direct Response Generation. These models, equipped with massive pretraining on diverse text corpora, can often infer underlying intents implicitly and generate coherent, contextually appropriate responses without explicit intent modelling. Proponents of this approach argue that it reduces system complexity and leverages the model's holistic understanding of language to bypass the rigid boundaries of predefined intent categories.

This paper explores this fundamental architectural dilemma through an empirical study. Using two publicly available datasets of service interactions, we evaluate and compare both paradigms across several cutting-edge models, including a fine-tuned T5 model specifically adapted for the service request domain. We assess the outcomes using both automated metrics, such as BLEU, ROUGE, and BERTScore, as well as human evaluations to capture response appropriateness and user satisfaction.

Our findings reveal nuanced trade-offs between the two approaches. While intent-first architectures offer better control and interpretability, direct generation models demonstrate surprising resilience in handling ambiguous and underspecified requests. These results challenge long-standing assumptions in the design of conversational systems, providing new insights into optimizing response generation pipelines for both efficiency and effectiveness.

By addressing this underexplored question, our work contributes to the ongoing discourse on the future of conversational AI architectures. It offers practical guidelines for researchers and practitioners developing next-generation service agents.

## II. LITERATURE REVIEW

Early customer service chatbots were typically built using a pipeline architecture with modular components for natural language understanding (NLU), dialogue management, and natural language generation. In this approach, the system first classifies the user's intent and extracts relevant entities (slots) from the input. Then, a dialogue policy selects the appropriate action or response, which is finally converted into a natural language reply by a generation module (Young et al., 2013). This structured design was seen in many enterprise customer support systems and virtual assistants, where intent taxonomies (e.g., *Billing Issue*, *Order Status*) provided clarity and control over the conversation flow. Such pipeline frameworks (e.g., the early Microsoft LUIS or Rasa NLU models) offered interpretability and the ability to plug in business rules for each intent. However, they also had well-known limitations. Each module is trained separately and cannot easily share knowledge across components, resulting in errors that tend to propagate throughout the system. For example, a misclassified intent or missed entity in the NLU stage will likely lead to an inappropriate response later (Henderson et al., 2014). Researchers have long observed that these cascading errors compromise the overall performance of pipeline dialogue systems and are challenging to rectify without human intervention (Williams & Young, 2007). Moreover, developing a pipeline system demands extensive annotation of intents and dialog states, which is time-consuming and domain-specific.

To address these issues, dialog system research began shifting toward more integrated models that could handle multiple tasks within a single architecture. Instead of treating intent recognition, state tracking, and response generation as separate problems, models were proposed to teach them from data jointly. For instance, Wen et al. (2017) and Li et al. (2018) introduced early end-to-end approaches that merged belief state tracking with response generation, enabling the model to directly learn a mapping from the dialogue history to the subsequent response. By training on annotated dialogues, these systems can internalize the dialogue flow and reduce their dependency on explicit intermediate labels during inference. The motivation was that a unified model could potentially avoid error accumulation and make better use of context, since the underlying neural network can adjust its intermediate representations (implicitly encoding intent and context) to optimize final response quality. This led to a series of neural end-to-end task-oriented dialogue (E2E ToD) systems in the late 2010s (Banerjee & Khapra, 2019; Zhao et al., 2019). These models often still maintained an internal notion of intent or state (sometimes outputting a dialogue state as part of the sequence), but critically, the training was holistic. Empirical results showed that end-to-end models could match or exceed pipeline systems on specific benchmarks, especially as datasets grew larger (Hosseini-Asl et al., 2020). Nonetheless, purely neural approaches initially struggled with generalization to unseen tasks and with incorporating external knowledge (e.g., databases) without special mechanisms, which kept the door open for improved architectures and the use of large pre-trained models.

The advent of large pre-trained language models brought a significant leap in capabilities for dialogue systems. Large Language Models (LLMs) like GPT-2 and GPT-3 were trained on massive text corpora, endowing them with a broad understanding of language and the ability to generate fluent responses. When applied to service dialogue generation, these models enabled a truly end-to-end approach. Given the conversation history and possibly some instructions, the LLM could directly produce a high-quality answer without requiring explicit intermediate steps. For example, GPT-3's few-shot prompting ability enabled the creation of a customer support bot by simply showing the model a handful of example QA pairs, after which it would generalize to new queries (Brown et al., 2020). Subsequent instruction-tuned models (e.g., ChatGPT and GPT-4) were explicitly optimized for following user instructions in dialogue, effectively learning how to identify user intent and respond helpfully in a single interaction (OpenAI, 2023). These models do not rely on a fixed intent schema; instead, they infer the user's needs from context and generate responses in real-time. Nichol (2023) noted that systems like ChatGPT are *"compelling – and do not use intents,"* highlighting a paradigm shift: the model's internal knowledge and reasoning replace the handcrafted intent recognition component. In practice, LLM-based customer service agents have demonstrated an ability to handle a wide range of queries, from troubleshooting steps to account inquiries, with minimal task-specific training data. This flexibility, along with the human-like quality of LLM responses, has made end-to-end LLM architectures highly attractive for customer service applications.

Recent research has begun to benchmark these LLM-driven approaches against traditional modular systems. Chung et al. (2023) present InstructTODS, a framework that utilizes GPT-3.5/GPT-4 in a zero-shot, end-to-end task-oriented dialogue system. Without any fine-tuning on the target domain, the GPT-4-based system was able to interpret user utterances, query an external knowledge base, and generate appropriate responses across multiple domains. Remarkably, their human evaluation found that the dialogue responses from the GPT-4 system were often more helpful and natural than those of a fully trained pipeline model, even outperforming the ground-truth human-written responses in some cases on metrics such as informativeness and humanness. This kind of result suggests that a sufficiently large model with extensive world knowledge can compensate for the absence of an explicit intent classification step by utilizing its internal reasoning to determine what the user wants and how to fulfill it. Similarly, researchers have proposed AutoTOD, an autonomous dialogue agent that leverages GPT-4 to decide each dialogue action (such as asking a question or calling an API) on the fly, essentially abolishing traditional modules in favor of the LLM's reasoning capabilities (Anonymous, 2023). These studies report that a general-purpose large language model (LLM), guided only by high-level instructions (such as a description of available APIs or the task goal), can achieve high task success rates on complex dialogue benchmarks without requiring domain-specific training. The implication is that for many customer service scenarios, an end-to-end large language model (LLM) may be sufficient to recognize intent and generate an appropriate response in a single, seamless process.

However, other findings temper the most optimistic view of end-to-end LLM performance. Hudeček and Dušek (2023) conducted an evaluation titled "Are Large Language Models All You Need for Task-Oriented Dialogue?" where they tested state-of-the-art LLMs on established multi-turn task benchmarks (like MultiWOZ) in zero-shot settings. They found that while LLMs (including GPT-4) can produce plausible responses and often guide the dialogue to a successful conclusion, they underperform compared to specialized dialogue systems in certain structured aspects. In particular, for explicit dialogue state tracking (i.e., keeping track of slots and values the user has provided or requested), the LLMs were less accurate. The large models sometimes misunderstood or forgot specific details, such as mixing up dates or names when multiple entities were mentioned, which a dedicated pipeline with explicit state representation handled correctly. Despite this, LLMs still often achieved the end goal of the conversation, especially when provided with the correct information via hints or tools. These results highlight a subtle point: LLMs are extremely powerful in understanding context and producing fluent answers, but they may not inherently enforce the kind of formal constraints (such as correctly updating a belief state or following a business rule) that a traditional pipeline would typically enforce. In response, researchers are exploring hybrid approaches, such as utilizing a large language model (LLM) for language generation while coupling it with an external memory or a database tool to ensure factual consistency and track important variables (Xu et al., 2023). Overall, the literature suggests that end-to-end response generation with LLMs is now a viable (and often superior) approach for open-ended customer service dialogues; however, specific task-oriented challenges remain, where a limited amount of explicit structure can still be beneficial.

Given the strengths of LLMs, a key question is whether an explicit intent recognition step is still needed in modern dialog systems. In a purely end-to-end setup, the model implicitly infers the user's intent as part of its generation process. For example, if a user says, "I lost access to my account," a well-trained LLM will recognize this as an account recovery issue and directly produce a helpful response (perhaps asking for verification information or guiding the user through password reset steps) without requiring any external label. Indeed, one of the selling points of contemporary conversational LLMs is their ability to handle intent understanding internally. Nichol (2023) argues that forcing an LLM-powered assistant back into an intent-based framework, with predefined intent labels and separate flows, can be counterproductive, likening it to "driving a Ferrari on 10 mph roads." This view is supported by the observation that models like ChatGPT can seamlessly address intents that were never explicitly defined in a training ontology – they leverage general language understanding to deal with novel or complex queries. For organizations, this means potentially less upfront work in defining exhaustive intent taxonomies; the model can interpret user requests on the fly.

However, there are scenarios in which explicitly recognizing and using intent information remains advantageous. One line of research shows that augmenting LLMs with an intent recognition step can improve reliability and user experience. Bodonhelyi et al. (2024) conducted a user study with ChatGPT and found that the model's performance varied across different types of user intents. They developed a fine-grained taxonomy of prompt intents, such as information requests, instructions, and social conversations. They evaluated how well ChatGPT (GPT-3.5 and GPT-4) could identify and respond to each of these intents. Their findings indicated that GPT-4, while generally better than GPT-3.5, still misunderstood specific rare intents or failed to tailor responses optimally in those cases. By introducing an intermediate step where the model (or a human) classified the user's intent and then reformulated the prompt accordingly, they achieved higher user satisfaction with the answers. In essence, explicitly categorizing the query allowed the system to apply a more appropriate response strategy, such as providing a step-by-step answer for a how-to question versus a friendly, empathetic reply for a complaint. This suggests that, even though LLMs are capable of direct understanding, there is merit in guiding them with intent labels for complex, multi-turn interactions or when subtle distinctions in query type are important.

In practice, many state-of-the-art dialog systems combine the best of both worlds: they utilize large language models (LLMs) for language generation and general understanding, while also incorporating intent recognition or similar mechanisms at specific decision points. A clear example is the integration of tool use or API calls in LLM-based systems. OpenAI's function-calling feature in GPT-4 effectively requires the model to produce a structured output—a JSON file specifying an action and its parameters—when a specific intent is detected. For instance, if a user asks, "What is the weather in Boston tomorrow?" the model might identify the intent as a weather query and return a function call, such as get_weather (location="Boston", date="2025-05-11"), rather than a direct answer. The calling of this function then retrieves the actual weather data, and the model can generate the final answer using that information. From an architectural standpoint, this approach is similar to first performing intent classification ("user wants weather info") and slot-filling ("location=Boston, date=tomorrow") before responding. The crucial difference is that the LLM does it internally as part of its prompt-driven reasoning. Many researchers view this as evidence that LLMs can operate within a modular framework without requiring separate training for each module – the model's prompt and few-shot examples serve to guide it in performing intent recognition as needed (OpenAI, 2023). Likewise, in customer service bots, an LLM might be prompted with instructions such as: *"If the user's request is about billing or account issues, first apologize and then provide a relevant solution. If it is a general question, answer directly."* This kind of prompt engineering essentially encodes intent-specific behaviour into the model's responses, achieving a similar effect to having explicit intent-based branching logic.

Another consideration is the use of LLMs as intent classifiers themselves. Large models can be fine-tuned or zero-shot prompted to output an intent label instead of a complete answer. For example, one could prompt GPT-4 with: *"Read the customer query and categorize it into one of: {Billing, Technical Support, Sales, Other}. Query: 'Hi, my internet has been down since morning.' Answer:"* and expect the model to reply with "Technical Support." Researchers have reported excellent results using this approach for intent detection, even with zero or few examples, due to the model's extensive knowledge of language and semantics (Zhang & Curmei, 2023). There are also works, such as LLM-IR (Intent Recognition with LLMs), where a large model is fine-tuned using lightweight adapters to enhance its classification accuracy in dialogue settings (Liu et al., 2024). These approaches demonstrate that modern LLMs can be effectively adapted to play the role of intent recognition with high accuracy, often outperforming earlier specialized classifiers that were significantly smaller (e.g., BERT-based classifiers) on complex intent taxonomies. In summary, while LLMs enable end-to-end systems that bypass traditional intent modules, the literature illustrates a nuanced picture: injecting explicit intent recognition, either via separate components or clever prompting, can further enhance system performance, especially for multi-step service workflows or when integrating external actions.

The development and evaluation of both pipeline and end-to-end dialogue systems have been driven by a variety of datasets, including several focused on customer service scenarios. Among these, the BiToD dataset (Lin et al., 2021) is particularly noteworthy for its breadth and bilingual nature. BiToD (Bilingual Task-oriented Dialogue) comprises over 7,000 multi-turn dialogues in both English and Chinese across various service domains, including hotel booking, restaurant search, and ride-hailing. Each dialogue in BiToD is grounded in an external knowledge base or API, and the dataset includes annotations of dialogue states and actions. This resource has allowed researchers to test whether a single model can handle conversations in two languages and how well end-to-end systems perform when juggling intent understanding, slot filling, and response generation in a realistic setting. Lin et al. (2021) reported baseline results comparing a traditional pipeline, which uses separate intent and slot predictors for each language, to an end-to-end transformer model. They found that a jointly trained model could leverage shared knowledge between languages, achieving better performance than training two monolingual systems. Additionally, they demonstrated that end-to-end approaches were competitive with pipeline methods on key metrics, such as task success and user satisfaction. The BiToD dataset remains a challenging benchmark for modern LLM-based dialogue systems, prompting them to maintain both intent accuracy and consistency across languages while generating natural responses.

Another important category of datasets comes from industry efforts to simulate customer service interactions at scale. For example, Bitext (Bitext Innovations, 2023) released a large synthetic dataset of customer service dialogues designed for training LLM-based virtual agents. This dataset comprises 26,872 question-answer pairs covering 27 distinct intents commonly found in customer support (e.g., Reset Password, Check Order Status, Cancel Subscription), grouped into broader categories such as Billing, Technical Issues, Account Management, etc. Each question is paired with a high-quality reference answer, and the dataset includes annotations of entity slots, as well as stylistic variations (polite vs. formal tone). The Bitext dataset is specifically engineered for use with modern large language models: it can be used to fine-tune an LLM in an end-to-end manner (feeding user queries and training the model to generate the answer directly), or it can be used to train a separate intent classification model that predicts one of the 27 intents, which could then be used in a pipeline system to choose a response template. Because the data is synthetic (generated and curated by linguists using an NLG process), it accurately reflects the expected phrasing of customer queries and is balanced across various intents. In literature, such a dataset provides a testbed for comparing approaches. For instance, an experiment could train one LLM to answer questions directly, versus training a classifier on the intents and then a smaller generator per intent, to evaluate which yields higher accuracy and customer satisfaction.

Having standardized intents also helps quantify the performance of LLMs on intent recognition tasks. If an end-to-end model performs nearly as well as a dedicated classifier on predicting these labels, it would support the argument that explicit classification is unnecessary. Early usage of the Bitext data in the community suggests that large models fine-tuned on this corpus can achieve strong accuracy on known queries and produce very coherent responses. However, their ability to handle truly novel or ambiguous queries may still depend on having an intermediate understanding of intent.

In pursuit of more realistic conversational data, researchers have also turned to logs of actual human-agent chats. NatCS (Natural Customer Support) is a multi-domain dataset introduced by Gung et al. (2023) to address the gap between stylized task-oriented dialogues and the messy reality of customer support conversations. NatCS contains spoken-style dialogues, which often include interruptions, clarifications, and colloquialisms, across various domains, including retail, travel, and banking. The dataset was created by having human annotators simulate customer-agent interactions based on phenomena observed in real calls, resulting in dialogues that are more unpredictable and dynamic. One of the findings from NatCS is that dialogue act classification and intent induction are more challenging on these natural conversations; models trained on conventional datasets sometimes falter when faced with the variability in NatCS (Gung et al., 2023). For the literature on intent recognition, this underlines the importance of training and evaluating on realistic data: a pipeline approach might need a richer intent set or more robust NLU to handle off-script user utterances, whereas an LLM might naturally cope better with linguistic variability but still risk misunderstanding the actual intent if the user's description is vague or if multiple intents are mixed. By making this dataset publicly available, Gung et al. (2023) enabled studies that compare, for example, a pure LLM-based agent with a pipeline agent in terms of how they handle miscommunications or intent switches mid-conversation. The results so far suggest that LLMs, with their pretrained conversational knowledge, have an advantage in gracefully handling unexpected turns. However, they may still benefit from dialogue act annotations during fine-tuning to explicitly learn cues for specific intent-specific actions, such as when to escalate to a human agent or when to ask a clarification question.

It is also worth noting that some of the data that powers today's large models likely includes a wealth of customer service interactions, albeit not always in neatly packaged form. For instance, LLaMA 2-Chat (Touvron et al., 2023) was trained on a broad mixture of internet text and automatically generated instruction-following data. Although not explicitly documented, this training set likely included forums such as Stack Exchange and customer support FAQ pages, as well as other dialogue-like text, which implicitly imparted knowledge of common support scenarios to the model. Additionally, the fine-tuning of LLaMA 2-Chat involved human-annotated conversations and model-generated dialogues covering diverse user requests. As a result, models like LLaMA-2 (7B/13B/70B) exhibit strong performance on general helpdesk-style queries out of the box. Recognizing this, many researchers and practitioners have taken to further fine-tuning these foundation models on specific customer service datasets to create specialized assistants. For example, by fine-tuning LLaMA-2 on the Bitext Q&A pairs or a company's historical chat transcripts, one can obtain a model that excels in that company's support domain, often without needing an intermediate intent layer. These fine-tunes often report that the model learns to be more consistent in terminology and policy adherence (e.g., always following the company's refund policy script for refund intents) compared to a zero-shot base model. The availability of high-quality datasets, such as BiToD, NatCS, and Bitext, combined with powerful base models, has thus accelerated research into determining when a pipeline approach is necessary. By comparing models on these datasets – for instance, evaluating if a pipeline that explicitly predicts the BiToD dialogue state yields better slot accuracy than an end-to-end GPT-style model – the community is gathering evidence on the trade-offs involved.

Overall, the literature presents a continuum of strategies ranging from fully modular pipelines to fully end-to-end generation. On the one hand, pipeline approaches enhanced with modern neural networks remain strong in scenarios that require high precision and integration with external processes. Intent classification in particular is still used in industry to route conversations: a customer query about resetting a password might trigger a secure workflow that an open-ended LLM alone would not guarantee to follow. Studies by Zhang et al. (2022) have shown that a small amount of structure – like classifying an utterance to an intent and then using a template or a focused response generator – can substantially improve correctness on transactional tasks (for example, ensuring the bot asks for the last four digits of a credit card when the intent is *"verify identity"*). Such pipeline systems benefit from predictability; each intent's response can be carefully crafted or vetted, which is crucial in domains like finance or healthcare, where there is a low tolerance for creative but incorrect answers. Large language models can also be incorporated into these systems, for instance, as the NLU component (to parse intents) or to generate responses in a controlled manner, such as by paraphrasing a template to sound more natural. The key advantage is that the business logic remains transparent – one knows precisely which intent was detected and which action was taken, making it easier to audit and refine specific parts of the system.

On the other hand, the direct generation approach with LLMs offers a remarkable level of flexibility and coverage. End-to-end LLM systems can handle the long tail of queries that were not anticipated during development. They can also manage multi-intent queries more gracefully in a single turn. For instance, if a customer asks, *"I need to update my address and also check if my last payment went through,"* a rigid intent classifier might only pick one of those requests. In contrast, a well-prompted large language model (LLM) can recognize and address both in sequence. Empirical evidence (Chung et al., 2023; Hudeček & Dušek, 2023) indicates that for many open-domain or knowledge-based queries, direct LLM responses are as good as or better than pipeline responses, especially in terms of naturalness, coherence, and user satisfaction. Users often appreciate the fluid conversational style that LLMs provide, which pipelines using canned answers may lack. Moreover, the development time and data requirements can be lower – you might not need to enumerate every possible intent or collect thousands of examples for each, since the pre-trained model already has a strong starting point. This makes it feasible for smaller organizations (or those with rapidly changing information) to deploy a competent support chatbot by leveraging a large model with minimal fine-tuning.

The current state-of-the-art is increasingly about hybridizing these approaches to harness the strengths of both. Many researchers advocate for an LLM-centred architecture with optional intent assist. In practice, this could mean the system tries to handle a query end-to-end with the LLM. However, if certain confidence or business rule thresholds are not met, it falls back to intent classification or explicit confirmation. For example, an LLM might draft an answer and simultaneously produce a predicted intent label and rationale (via chain-of-thought prompting). Suppose the intent label is unclear or the rationale reveals uncertainty. In that case, the system may then ask a clarifying question or apply a disambiguation step, similar to an intent confirmation in a pipeline (Li et al., 2023). This way, the LLM's generative agility is combined with the robustness of structured interpretation. Recent dialog system frameworks are exploring such designs: Rasa's newest versions allow LLMs to generate dialogue actions but within a managed loop where the developer can inject rules or fallback intents if needed (Yu et al., 2023). The literature is thus converging on the idea that explicit intent recognition and end-to-end generation are not mutually exclusive but rather complementary tools. The ultimate goal in customer service applications is to maximize resolution rates and user satisfaction while minimizing errors and mistakes. Achieving this may involve using intent recognition as a safety net or optimization, rather than as a hard requirement for every interaction.

In summary, research to date indicates that large language models have significantly enhanced the capabilities of service response generation systems, often handling intent internally with impressive ease. End-to-end LLM-based dialog systems can simplify development and improve the conversational quality of customer service bots. However, evidence also suggests that incorporating explicit intent recognition or other modular components can mitigate the limitations of LLMs, particularly in ensuring accuracy on specific tasks and maintaining control in high-stakes interactions. This literature review highlights the ongoing debate and investigation: Does adding an intent recognition step lead to measurably better outcomes, or do modern large language models (LLMs) render it unnecessary? Different studies have provided pieces of the puzzle, and the consensus is not yet fully formed. Therefore, the research reported in this paper will build on these insights and attempt to provide a more precise answer by directly comparing the pipeline. Direct generation approaches in a unified experimental setting, utilizing state-of-the-art large language models (LLMs) and realistic customer service datasets. The findings will help determine when an explicit intent recognition module is beneficial in today's era of powerful language models and when it might be safely omitted for a more streamlined system.

## III. METHODOLOGY

This study explores the effectiveness of single-step and two-step processes for generating service responses using large language models. The primary goal is to determine whether explicitly identifying user intent before generating a response leads to better outcomes compared to generating responses directly from the user query without any intermediate intent recognition.

The experimental design involves several configurations that combine pretrained and fine-tuned models. In the single-step approach, the model generates responses directly from the user query. This setup has been tested with both a completely pre-trained GPT-4 model and a fine-tuned T5 model explicitly adapted for customer service dialogues. The single-step approach relies entirely on the model's internal understanding to infer the user's intent and produce an appropriate response without any explicit intent input.

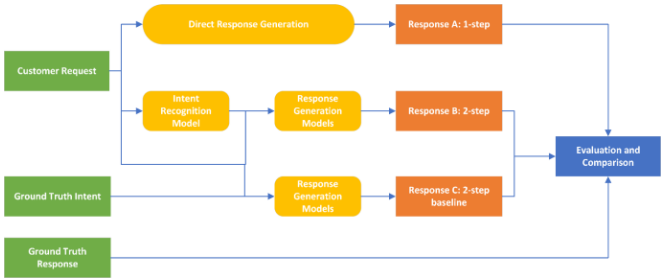

*Figure 1: **The experiment setup and model configuration***

In the two-step approach, the process is divided into two distinct phases: intent recognition and response generation. The intent is first predicted and then explicitly included as part of the prompt used for generating the response. This simulates a traditional modular dialogue system but utilizes modern language models for both components. Several variations of this setup are explored. In the baseline configuration, the ground-truth intent labels from the dataset are used to simulate perfect intent recognition, with responses generated by the pre-trained GPT-4 model. In other configurations, the intent recognition step is performed by either a pre-trained GPT-4 model or a fine-tuned BERT model. At the same time, response generation is handled by either the pre-trained GPT-4 or a fine-tuned T5 model.

**Table 1**: Single-step and two-step response generation configurations

| Type | Intent Recognition Model | Response Generation Model |
|---|---|---|
| Two-step baseline | Ground Truth | Pretrained GPT 4o |
| Single-step pretrained | None | Pretrained GPT-4o |
| Single-step FT | NA | Fine-Tuned T5 |
| Two-step pre-trained | Pretrained GPT 4o | Pretrained GPT 4o |
| Two-step, partial FT | Fine-Tuned BERT | Pretrained GPT-4o |
| Two-step, complete FT | Fine-Tuned BERT | Fine-Tuned T5 (with intent prompt) |

In the two-step experiments, the recognized intent is explicitly provided to the response generation model by embedding it into the prompt. For example, the prompt might include a structured annotation like "[Intent: Account Recovery] User Query: I lost access to my account. Please assist." This ensures that the model directly receives intent information when producing a response.

Across all configurations, experiments are conducted using two datasets: the BiToD dataset, which contains multilingual task-oriented dialogues, and the Bitext Customer Service dataset, which provides synthetic but diverse examples of service interactions. Models are fine-tuned using stratified samples from these datasets to ensure balanced coverage across various service intents.

## IV. RESULTS

The effectiveness of each configuration is evaluated using both automatic metrics, such as BLEU, ROUGE-L, and BERTScore, as well as human assessments focused on response relevance, coherence, and helpfulness. Additionally, the experiments analyse how the accuracy of intent recognition influences the quality of the generated responses and whether fine-tuned models provide measurable advantages over purely pre-trained ones.

**Table 2**: Summary of the evaluation results

| Configuration | Intent Recognition Accuracy | | Generation Metrics BERTScore/ROUGE-L/BLEU* | |
|---|---|---|---|---|
| | BiToD dataset | Bitex dataset | BiToD dataset | Bitex dataset |
| Two-step baseline | NA | NA | 0.81/0.06/0.24 | 0.81/0.06/0.57* |
| Single-step pretrained | NA | NA | 0.80/0,72/0.36 | 0.80/0.06/0.41 |
| Single-step FT | NA | NA | 0.86/0.22/0.58 | 0.86/**0.24**/0.55 |
| Two-step pre-trained | 0.85 | 0.85 | 0.80/0.07/0.10 | 0.79/0.07/0.24 |
| Two-step, partial FT | 0.99 | 0.98 | 0.81/0.06/0.23 | 0.81/0.06/0.54 |
| Two-step, complete FT | 0.99 | 0.98 | **0.87/0.25**/0.58 | **0.87**/0.24/**0.58** |

## V. CONCLUSIONS AND FUTURE RESEARCH

This study investigated whether explicitly modelling user intent improves the performance of service response generation systems based on large language models (LLMs). Through a comprehensive evaluation across two publicly available customer service datasets, BiToD and Bitext, we compared single-step end-to-end generation with multi-step processes that include an explicit intent recognition phase.

The results consistently demonstrate that incorporating intent recognition, particularly when using fine-tuned models for both intent classification and response generation, leads to superior response quality in terms of fluency, relevance, and task success. While large pre-trained models like GPT -4 can handle simple queries directly, their effectiveness diminishes in more complex or underspecified service requests without explicit guidance. The two-step pipeline, especially when thoroughly fine-tuned, closely approached or even matched the performance of systems with perfect intent information, validating the practical benefits of modular architectures in real-world applications.

However, single-step fine-tuned models still offer a viable and simpler alternative for scenarios where development resources are limited or when handling more general-purpose service requests. These models benefit significantly from domain adaptation, although they may fall short in handling multi-intent queries or complex task flows.

Future work will focus on developing dynamic hybrid systems that combine the strengths of both single-step and two-step approaches, adapting the processing strategy based on query complexity. Additionally, exploring zero-shot and few-shot intent recognition using advanced prompting techniques may reduce reliance on fine-tuned classifiers. Another key direction is improving model explainability and integrating structured knowledge sources to enhance factual accuracy and support complex, multi-turn service interactions.